\documentclass[letterpaper]{article} % DO NOT CHANGE THIS
\usepackage{aaai2026}  % DO NOT CHANGE THIS
\usepackage{times}  % DO NOT CHANGE THIS
\usepackage{helvet}  % DO NOT CHANGE THIS
\usepackage{courier}  % DO NOT CHANGE THIS
\usepackage[hyphens]{url}  % DO NOT CHANGE THIS
\usepackage{graphicx} % DO NOT CHANGE THIS
\usepackage{natbib}  % DO NOT CHANGE THIS AND DO NOT ADD ANY OPTIONS TO IT
\usepackage{caption} % DO NOT CHANGE THIS AND DO NOT ADD ANY OPTIONS TO IT
\usepackage{newfloat}
\usepackage{listings}
\DeclareCaptionStyle{ruled}{labelfont=normalfont,labelsep=colon,strut=off} % DO NOT CHANGE THIS
\title{Context-Sensitive Abstractions for \\Reinforcement Learning with Parameterized Actions}
\author{
    Rashmeet Kaur Nayyar\equalcontrib$^1$,
    Naman Shah\equalcontrib$^{1,2}$, and
    Siddharth Srivastava$^{1}$
}
\affiliations{
    \textsuperscript{\rm 1}Arizona State University, Tempe, AZ, USA\\
    \textsuperscript{\rm 2} Brown Unviersity, Providence, RI, USA\\
    \{rmnayyar, shah.naman, siddharths\}@asu.edu
}

\usepackage{bibentry}
\usepackage{graphicx}
\usepackage{algorithmic}
\usepackage[linesnumbered,ruled,vlined]{algorithm2e}
\usepackage{amsthm}
\usepackage{amsmath}
\usepackage{amssymb}

\theoremstyle{definition}
\newtheorem{definition}{Definition}[section]
\usepackage{amsmath,amsfonts,bm}

\def\eqref#1{equation~\ref{#1}}
\def\1{\bm{1}}

\DeclareMathAlphabet{\mathsfit}{\encodingdefault}{\sfdefault}{m}{sl}
\SetMathAlphabet{\mathsfit}{bold}{\encodingdefault}{\sfdefault}{bx}{n}

\def\gA{{\mathcal{A}}}

\def\gC{{\mathcal{C}}}
\def\gD{{\mathcal{D}}}
\def\gE{{\mathcal{E}}}

\def\gK{{\mathcal{K}}}
\def\gL{{\mathcal{L}}}
\def\gM{{\mathcal{M}}}
\def\gN{{\mathcal{N}}}

\def\gP{{\mathcal{P}}}

\def\gS{{\mathcal{S}}}

\def\gV{{\mathcal{V}}}

\DeclareMathOperator*{\argmax}{arg\,max}

\newcommand{\alg}{PEARL}
\newcommand{\spacatname}{SPA-CAT}

\newcommand{\spacat}{\Delta}

\newcommand{\cat}{\Delta}

\newcommand{\Sm}{\mathcal{S}}

\newcommand{\absS}{\overline{\mathcal{S}}}

\newcommand{\abss}{\overline{s}}
\newcommand{\absA}{\overline{\mathcal{A}}}

\newcommand{\absa}{\overline{a}}

\usepackage{float}
\usepackage{subcaption}
\usepackage{booktabs}
\usepackage[table]{xcolor}
\usepackage{mathtools}

\begin{document}

 \maketitle

\begin{abstract}
Real-world sequential decision-making often involves parameterized action spaces that require both, decisions regarding discrete actions and decisions about continuous action parameters governing how an action is executed. Existing approaches exhibit severe limitations in this setting---planning methods demand hand-crafted action models, and standard reinforcement learning (RL) algorithms are designed for either discrete or continuous actions but not both, and the few RL methods that handle parameterized actions typically rely on domain-specific engineering and fail to exploit the latent structure of these spaces. This paper extends the scope of RL algorithms to \emph{long-horizon}, \emph{sparse-reward} settings with parameterized actions by enabling agents to autonomously learn both state and action abstractions online. We introduce algorithms that progressively refine these abstractions during learning, increasing 
fine-grained detail in the critical regions of the state–action space where greater resolution improves performance.
Across several continuous-state, parameterized-action domains, our abstraction-driven approach enables $TD(\lambda)$ to achieve markedly higher sample efficiency than state-of-the-art baselines.

% Empirical results show that learning such abstractions on-the-fly enables $TD(\lambda)$ to substantially outperform state-of-the-art RL methods in terms of sample efficiency across diverse domains with continuous states, long horizons, and parameterized actions.

% Real-world sequential decision making problems often require parameterized action spaces that require both, decisions regarding discrete actions and decisions about continuous action parameters governing how an action is executed. However, existing approaches exhibit severe limitations when handling such parameterized action spaces---planning algorithms require hand-crafted action models, and reinforcement learning (RL) paradigms focus on either discrete or continuous actions but not both. This paper extends the scope of RL algorithms to long-horizon, sparse-reward settings with parameterized actions through autonomously learned state and action abstractions. We present algorithms for online learning and flexible refinement of such abstractions during RL. Empirical results show that learning such abstractions on-the-fly enable $TD(\lambda)$ to significantly outperform state-of-the-art RL approaches in terms of sample efficiency across diverse problem domains with long horizons, continuous states, and parameterized actions.
\end{abstract}

\begin{links}
% \textcolor{red}{read abstract}
    \link{Code}{https://github.com/AAIR-lab/PEARL.git}
    \link{Extended version}{https://aair-lab.github.io/Publications/nss-aaai26.pdf}
\end{links}

\section{Introduction}
Reinforcement learning (RL) has delivered strong results across a diverse range of decision-making tasks, from discrete action settings like Atari games \citep{mnih2015human} to continuous control scenarios such as robotic manipulation \citep{schulman2017proximal}. 
Yet most leading RL approaches \citep{schulman2017proximal, haarnoja2018soft, schrittwieser2020mastering, hansen2023td}
are designed for either discrete or continuous action spaces---not both. Many real-world problems violate this dichotomy.
In autonomous driving, for example, the agent must choose among qualitatively distinct actions (accelerate, brake, turn), each endowed with discrete or continuous parameters such as braking force or steering angle. Such actions---known as \textit{parameterized actions}---require choosing not only the action but also determine its (real-valued) parameters before execution.

% In such cases, the agent must not only select an action but also determine its (real-valued) parameters prior to execution. 
% Such actions are known as \textit{parameterized actions}.

\begin{figure}[t!]
    \centering
        \includegraphics[width=0.75\columnwidth]{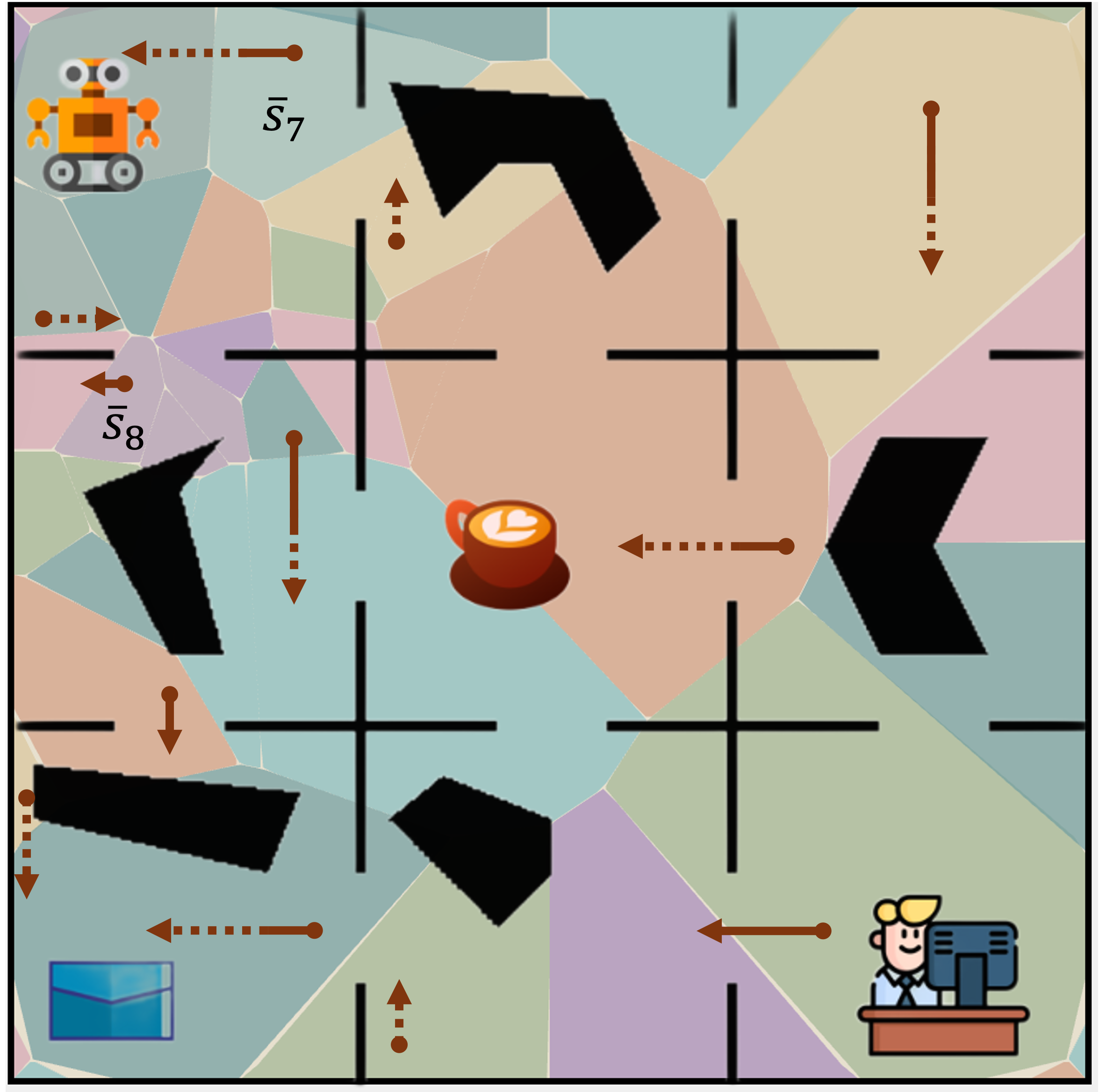}
        \caption{
        In a continuous version of the office domain, the agent needs to learn policies for delivering multiple items. Polygonal cells illustrate learned state abstractions, and arrows illustrate learned policies with abstract actions parameterized by parameter intervals. Each arrow corresponds to an interval $[a,b)$ of possible movement values:  the solid segment indicates the lower bound $a$, and the dotted segment indicates the interval width $b-a$. Narrower dotted segments denote higher precision in the learned action parameters.
        % Illustration of abstract states (shown as colored regions) and abstract actions for parameterized actions (shown as green dashed arrows representing parameter range; solid lines represent the minimum) in the OfficeWorld domain. 
        % In the tightly constrained space of abstract state $\abss_8$, the left action requires a finely tuned movement parameter (higher precision), whereas the more open layout of $\abss_7$ permits a far coarser parameter for navigation (lower precision).
}
        \label{fig:spacat1}
\end{figure}

While recent methods have made progress in addressing parameterized actions ~\citep{xiong2018parametrized, bester2019multi, li2021hyar}, they largely ignore 
utilizing
the underlying structure inherent in parameterized-action spaces. In navigation tasks, for instance, an agent should adjust movement parameters with high precision near obstacles but can act with much coarser control in open areas.
% E.g., in autonomous navigation, an agent might benefit from making fine-grained adjustments to movement parameters near obstacles, while only needing coarse-grained control in open, obstacle-free areas.
% while learning autonomous navigation policies one could consider fine-grained decisions over movement actions' parameters near obstacles, while considering only coarse partitions of parameter ranges  in wide empty spaces.
Existing approaches also often rely on carefully engineered dense rewards and environment-specific initializations to facilitate learning
% expert-designed dense reward functions 
or benefit from relatively short ``effective horizons'' to remain tractable~\citep{laidlaw2023bridging}.
% several approaches rely heavily on an expert-curated dense reward function aiding policy learning, or a relatively smaller ``effective horizon'' of the problem~\citep{laidlaw2023bridging}.
%SS: not sure what uai2023 ref was doing here.
% This significantly reduces the applicability of such approaches to real-world scenarios with long horizons and sparse rewards---settings where environment feedback is infrequent and delayed. 
% As a result, the challenge of learning and leveraging the structure of parameterized action spaces to improve sample efficiency, particularly in long-horizon tasks, remains largely unaddressed. 
% The problem of learning and utilizing  the structure of parameterized action spaces for sample efficiency in long-horizon problems remains understudied. 
A detailed discussion of related work is in Sec.~\ref{sec:related}.

This paper aims to extend the scope and sample efficiency of RL paradigms to relatively under-studied yet challenging class of problems that feature long horizons, sparse rewards, and parameterized actions. We introduce the first known approach called \alg{} that automatically discovers structure in parameterized-action problems in the form of conditional abstractions of their state spaces and action spaces. As an illustration, Fig.~\ref{fig:spacat1} shows flexible abstraction of the state space and how the policy may require a different extent of action abstraction in different states in the OfficeWorld domain: in the tightly constrained region $\abss_8$, navigation demands high-precision in action parameters, whereas the more open space of $\abss_7$ tolerates far coarser abstraction. This contrast highlights why abstractions must capture this variation in the required precision of action parameters across different regions of the state space.

% \textcolor{red}{Intuition of key idea of approach, Novelty diff with prior work}

Given an input problem in the RL setting where a state is expressed using discrete and continuous state variables and an action is expressed using continuous or discrete parameters, PEARL learns context-sensitive abstractions while performing $TD(\lambda)$. It uses a combination of dispersion in TD-error and value-function signals to learn which abstract states and action parameters require finer resolution during learning.
Our approach builds upon our recent work on conditional state abstractions \cite{dadvar2023conditional}, and introduces new algorithms for learning more general forms of  state abstractions alongside abstractions of action parameters.

% Our work builds upon recent developments in learning conditional state abstractions for RL~\cite{dadvar2023conditional}, and develops novel algorithms for learning abstractions of action parameters coherently with new approaches for learning  abstractions. 

Our main contributions are: (1) A unifying formal framework for context-sensitive abstractions of continuous state spaces and parameterized actions with continuous arguments; (2) an approach for learning flexible refinements of abstractions; (3) algorithms for learning such state and action abstractions on the fly, during RL, and thereby exploiting latent structural properties of problem instances for efficient learning without any hand-crafting of abstractions; (4) an evaluation of this approach as applied to $TD(\lambda)$, showing that using this abstraction paradigm with foundational RL paradigms improves their performance beyond state-of-the-art algorithms.

\section{Preliminaries}
% [1. define mdp
% 2. states
% 3. parameterized actions
% 4. transition function
% 5. policy
% 6. Running example
% 7. state abstraction
% 8. action abstraction]

We use the framework of episodic factored goal-oriented Markov decision process (MDP) with parameterized actions~\citep{bertsekas2011dynamic,
hausknecht2015deep,
deng2022polynomial}. An MDP $\gM$ is defined as $\langle \gV, \gS, \gA, T, R, \gamma, h, s_0, G \rangle$, where $\gV$ is a set of state variables and the domain of each variable $v \in \gV$ is a bounded interval $\gD_{v_i} = \left[ \gD^\emph{min}_{v_i}, \gD^\emph{max}_{v_i} \right] \subseteq \mathbb{R}$; 
% with $\gD^\emph{min}_{v_i}$ and $\gD^\emph{max}_{v_i}$ denoting the minimum and maximum values the variable can take respectively. 
$\gS$ denotes the set of factored states defined by $\gV$, where a state $s \in \gS$ is an assignment of values to all variables in $\gV$:
$s = \{ v_i = x_k | v_i \in \gV \land x_k \in \gD_{v_i} \}$. We use $s(v_i)$ to denote the value of variable $v_i$ in state $s$.

The action set $\gA$ consists of a finite number of stochastic parameterized actions.
% , each parameterized by continuous variables.
Each action $a \in \gA$ is a parameterized function $a_l(a_p)$, where $a_l$ is the action label and $a_p =\langle x_1,\dots,x_k \rangle$ is an ordered set of $k$ continuous parameters where each parameter $x_i$ has a bounded and ordered domain $\gD_{x_i} \subseteq \mathbb{R}$.  The complete parameter space is defined as $\gP_a = \bigtimes_{i=1}^k \gD_{x_i}$. A grounded action $\tilde{a_i}$ assigns values to these parameters from their respective domains. The set of all possible grounded actions is denoted $\tilde{\gA}$, and may be infinite given continuous parameters.

The transition function
$T: \gS \times \tilde{\gA} \rightarrow \mu\gS$
defines a distribution over next states, given a state and a grounded action. The reward function $R: \gS \times \tilde{\gA} \rightarrow \mathbb{R}$ assigns scalar rewards to state-action pairs. The discount factor $\gamma \in [0,1]$ determines the weights of future rewards, and $h$ is the episode horizon. $s_0$ is the initial state and $G$ is the set of goal states.

The objective is to learn a policy $\pi_\gM: \gS \rightarrow \tilde{\gA}$ that when executed from the initial state $s_0$, reaches a goal state in $s_g \in G$ while maximizing the expected cumulative discounted reward $\mathbb{E}_{\pi}[\sum_{t=0}^{t=h}\gamma ^t r_t]$. We use the RL setting, where both $T$ and $R$ are unknown~\citep{sutton1998introduction}. 

The state-value function \( V^\pi(s) \) under a policy \( \pi \) denotes the expected return starting from state \( s \) and following \( \pi \):
\[
V^\pi(s) = \mathbb{E}_\pi \left[ \sum_{t=0}^h \gamma^t r_t \mid s_0 = s \right]
\]
The action-value function \( Q^\pi(s, \tilde{a}) \) gives the expected return starting from state \( s \), executing action \( \tilde{a} \), and thereafter following \( \pi \):
\[
Q^\pi(s, \tilde{a}) = \mathbb{E}_\pi \left[ \sum_{t=0}^h \gamma^t r_t \mid s_0 = s, \tilde{a}_0 = \tilde{a} \right]
\]

\paragraph{TD($\lambda$)} We use TD($\lambda$)~\citep{sutton1988learning} for learning the policy $\pi$. It combines one-step TD and Monte Carlo methods by weighting updates across multiple future time steps, controlled by the trace-decay parameter $\lambda \in [0,1]$. 
% This method updates Q-value estimates after each step using accumulated traces from recent states and rewards.

\paragraph{Abstraction} Abstraction has been recognized as a key mechanism for achieving scalability in long horizon, sparse reward settings ~\citep{li2006towards, shah2024hierarchical,  wang2024building}.  A state abstraction is a mapping $\alpha:\Sm\rightarrow\absS$ that assigns each concrete state $s \in \Sm$ to an abstract state $\abss \in \absS$, where $\absS$ is a partitioning of the original state space $\Sm$. In this work, we define an analogous notion of abstraction for an action, defined as a partitioning of the action-parameter space (formalized in Sec.~\ref{sec:spa_cats}).

We now describe our approach for efficiently learning a policy in settings with parameterized actions by automatically learning context-sensitive state and action abstractions.

\section{Our Approach}

% - Overall contribution and idea of the approach
% - explain the core idea through an example

The central contribution of this paper is a novel abstraction paradigm for jointly representing and learning state and action abstractions.
%, and an approach for jointly learning these abstractions. 
%The central contribution of this paper is to devise the notion of abstract states and abstract actions 
These abstractions  exploit the structure of the environment
%, and an abstraction structure that provides a unified representation for state and action abstractions 
in order to efficiently learn and represent  policies for problems with parameterized actions.
% that achieves the posed goal. 

% - Running example: Consider the 
\paragraph{Running example} Consider an AI agent in an Office environment (Fig.~\ref{fig:spacat1}) that must collect and deliver a coffee and a mail between rooms and offices. The state variables include the agent's $(x, y)$ position with $x, y \in [ 0.0, 5.0)$, and two binary variables: $c \in \{0, 1\}$ and $m \in \{0, 1\}$ indicating whether it is carrying coffee or mail. 
The agent has four actions to move in the cardinal directions, i.e., $\gA = \{ \emph{up}(d), \emph{down}(d), \emph{left}(d), \emph{right}(d) \}$, each with one continuous parameter  $d \in [0, 0.5)$ that determines the movement distance. Actions may result in stochastic displacements along orthogonal directions, and the agent picks or drops items automatically at designated locations. This setting extends the OfficeWorld environment ~\citep{icarte2022reward} by incorporating parameterized actions. 

\begin{figure}[t]
    \centering
    \includegraphics[height=2.6in]{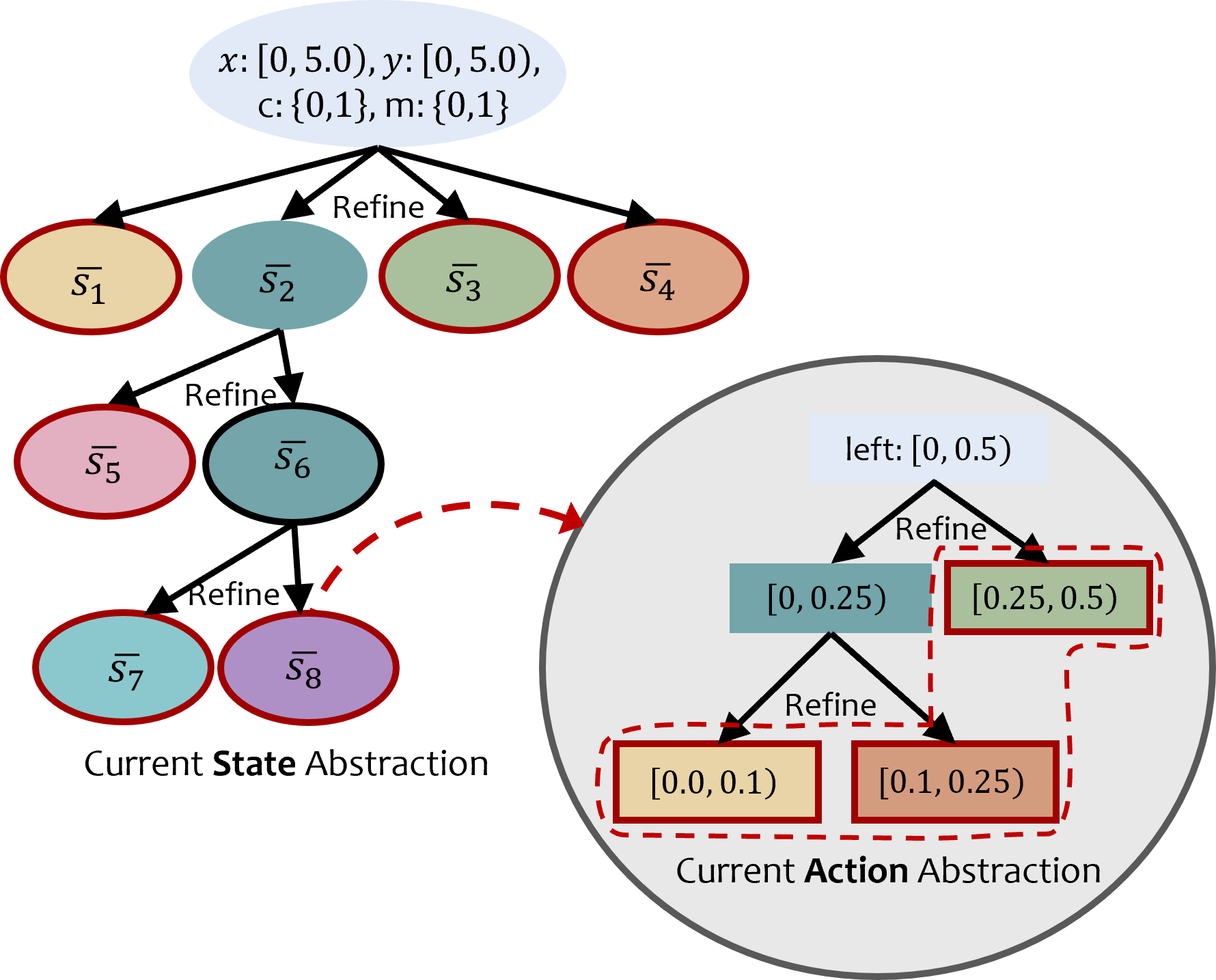}
    \caption{Illustration of a \spacatname{} for Office World.}
    \label{fig:spacat3}
\end{figure}

This work builds upon our prior work~\citep{dadvar2023conditional}
which learned state abstractions with strictly uniform refinements. It did not address the problem of parameterized actions and offered no mechanism for more flexible refinements of abstractions. The  abstraction learned in this work has the following desirable properties: (i) The abstractions are flexible---the abstract state boundaries are not constrained to be orthogonal or axis-aligned. This flexibility allows the learned abstractions to better adapt to the geometry and dynamics of the environment---for example, by placing boundaries where agent behavior changes, following the contours of obstacles. Fig.~\ref{fig:spacat1} illustrates an example of such a state abstraction for Office World, where each colored region represents a distinct abstract state. (ii) Moreover, each abstract state has a conjoined action parameter tree for each action (shown in Fig.~\ref{fig:spacat3}), allowing varying levels of precision in different abstract states. 
E.g., in open areas—such as the centers of rooms (e.g., abstract state $\abss_7$)—the agent can move freely without requiring high precision in selecting movement distances (e.g., abstract action \emph{left}([0.25,0.5))). In contrast, in more constrained areas—such as corridors, near obstacles, or narrow passages (e.g., abstract state $\abss_8$)—precise control over movement is crucial (e.g., abstract action \emph{left}([0.0,0.1))). Fig.~\ref{fig:spacat3} shows the unified state-action abstraction tree, where the leaves represent these abstract states and abstract actions. The shown abstractions capture the required higher precision for selecting action parameters. 
% tighter spaces, such as corridors, near obstacles, or narrow passages, demand greater precision in action selection.
% In Fig.~\ref{fig:spacat1}, $\abss_7$ and $\abss_8$ form distinct abstract states since they require different levels of precision in executing the same action (i.e., different abstract actions in Fig.~\ref{fig:spacat1}), due to the presence of nearby obstacles. 
We hypothesize that automatically identifying such meaningful abstract states and corresponding action parameter trees can significantly improve sample efficiency in policy learning.
% We postulate that if we identify such abstract states and actions automatically, such a structure allows sample-efficient policy learning.
 
We now define our unified framework for jointly representing both state and action abstractions.
% We associate an abstraction of the parameter space of an action with each abstract state. The next section formalizes these abstractions and describes our unified representation for jointly representing state and action abstractions. 

\subsection{State and Action Abstractions}
\label{sec:spa_cats}

This section formalizes our representations for state and action abstractions, beginning with action parameter abstractions, followed by an integrated representation for state and action abstractions. 
We use \emph{action parameter trees} (APTs) to formalize the intuitive example of action parameter abstractions discussed above. 
%APTs represent a hierarchical structure for representing action parameter abstractions.
%Intuitively, an APT represents parameter abstraction for a given action. 
Each node in an APT represents a %partition%
susbset of the parameter space of an action, and its children nodes together represent a partition of that subset. Formally, given a parameterized action $a \in \gA$ with a complete parameter space $\gP_a$, we define a corresponding APT as follows: 

\begin{definition} [Action Parameter Tree (APT)] An APT $\tau$ is a directed hierarchical structure defined as a tuple $\langle \gN, \gE, N_0, \ell \rangle$ where $\gN$ is a set of nodes, $\gE$ is a set of edges such that each $(u,v) \in \gE$ represents a directed edge from node $u$ to node $j$. 
%A node $N_i$ is a parent node of $N_j$ iff $(N_i, N_j) \in \gE$. Correspondingly, a node $N_j$ is a child node of $N_i$ iff $(N_i,N_j) \in \gE$. 
$N_0$ is the root node. $\ell: \gN \rightarrow 2^\gP$ defines a labeling function that maps each node $n_i \in \gN$ to a subset of the parameter space such that $\ell(N_0) = \gP$, the complete set of parameter values.
% set of possible parameter-tuple values. 
The set of all children nodes $\{n_j\}_{\{j=1,\ldots,k\}}$  of node $n_i$ represent a partition of $\ell(n_i)$, i.e, $\cup_{j=1,\dots,k}~~ \ell(n_j) = \ell(n_i)$ with labels of children nodes representing mutually exclusive sets.
%and $\cap_{j=1,\dots,k} \ell(N_j)=\emptyset$. 
\end{definition}

Given an APT $\tau_a$ for a parameterized action $a \in \gA$, we define $\gL_\tau \subseteq \gN_\tau$ 
as the set of leaf nodes or the ``fringe'' of $\tau_a$. 
The tree structure is learned autonomously so that at any stage of learning, the fringe of an action's APT represents the current abstraction of its parameter space. In this way, the fringe of $\tau_a$ can be used to define a set of abstractly grounded versions of $a$, where each version picks its parameters from one of the leaves of $\gL_\tau$.  

Formally, the set of abstract parameter sets defined by an APT $\tau$ for an action labeled $a$ is defined as
$\bar{\gA}_{\tau} = \{\ell(n) | n \in
%\gN_\tau \}$  %%%fixed error. double check
\gL_\tau \}$. Given a concrete action $a(q)$ and an APT $\tau$ for $a$, we use  $\bar{q}_{\tau}$ to denote the unique element of $\bar{\gA}_{\tau}$ that includes $q$. Thus, $a(\bar{q}_{\tau})$ denotes the abstraction of $a(q)$ under $\tau$. For brevity, we will use the  form $\bar{a}$ to denote an abstract version of $a$, and $\tilde{a}$ to denote its concrete grounded version (henceforth referred to as a ``concrete action'') with real-valued parameters.  During RL, we execute an abstract action $\bar{a}= a(\bar{q}_\tau)$ by sampling its parameters $q$ uniformly from  $\bar{q}_\tau$ to obtain a concrete executable action $\tilde{a} = a(q)$.

We define unified state and action abstractions using  state and parameterized-action conditional abstraction trees (SPA-CATs). Intuitively, each node of a SPA-CAT defines a subset of the state space and is associated with its own APTs for each action. The structure of the state abstraction part of the tree is congruent with our notion of APTs but applied to the state space. Formally,

\begin{definition}[State and Parameterized Action Conditional Abstraction Tree (SPA-CAT)]
    A \spacatname{} $\cat{}$ is a directed hierarchical structure defined as a tuple $\langle \gN, \gE, N_0, \ell_s, \ell_a,  \rangle$, where $\gN$ is a set of nodes, $\gE$ is a set of directed edges. Each edge $(u,v) \in \gE$ defines a directed edge between nodes $u$ to $v$. $N_0 \in \gN$ defines a root node without a parent node. $\ell_s: \gN \rightarrow 2^\gS$ defines a labeling function that maps each node $n \in \gN$ to a subset of the state space $\gS_n \subseteq \gS$ such that $\ell_s(N_0) = \gS$. The set of all children nodes  $\{n_j\}_{\{j=1,\dots,k\}}$ of a node $n_i$ repesent a partition of $\ell(n_i)$, i.e., $\cup_{j=1\dots,k}~\ell_s(n_j) = \ell_s(n_i)$, with labels of children nodes representing mutually exclusive sets. 
    %for all children nodes of a node $N_i$, $\{N_j\}_{\{j=1,\dots,k\}}$, $\cup_{j=1\dots,k}\ell(N_j) = \ell_s(N_i)$ and $\cap_{j=1,\dots,k}\ell(N_j)=\emptyset$.
    $\ell_a: \gN \times \gA \rightarrow \Theta$ maps node $n_i \in \gN$ and a parameterized action $a_j \in \gA$ to an APT $\tau_{a_j}$ in the set of all possible APTs, $\Theta$.
\end{definition}

SPA-CATs define state and action abstractions as follows.
The set of leaf nodes (or the ``fringe'') of a SPA-CAT $\Delta$,  denoted as $\gL_\Delta \subseteq \gN_\Delta$,  define an abstract state space: $\bar{\gS}_\Delta = \{ \ell_s(n) | n \in \gL_\Delta \}$. Let $n_\Delta(s)$ denote the unique fringe node of $\Delta$ that represents $s$. The \emph{abstraction of a concrete state $s$} under $\Delta$, $\bar{s}_\Delta$, is defined as the set represented by the unique fringe element that includes $s$:  $\bar{s}_\Delta=\ell_s(n_\Delta(s))$. Further, each node $n$ in the fringe is associated with an APT $\ell(n, a)$ for each $a\in \gA$. This allows us to define the abstraction of a grounded action $a(q)$ relative to a concrete state $s$ and a SPA-CAT $\Delta$, $\bar{a}_{s, \Delta}$, as $a(\bar{q}_\tau)$, where $\tau = \ell(n_\Delta(s), a)$. 
We omit subscripts when clear from context. In this representation, each abstract state defines its own APTs. This allows the agent to tune the level of precision in each action's abstraction as a function of the current state. This is particularly conducive for compact expressions of $Q(s,a)$ functions in RL.
%
% set of grounded abstract actions $\bar{\gA} = \{ (a, \ell_a(N_i,a) | N_i \in \gL \land a \in \gA \}$. 
We now discuss our approach for automatically learning SPA-CATs from scratch during RL.

% \subsection{Statistical Learning-Based Flexible Abstraction Refinement} 
\subsection{Learning Abstraction Trees}
\label{sec:refinement}
 Throughout this work, we use abstraction trees introduced above to express Q functions. In particular, we express and maintain an abstract  Q function as a mapping from the abstract states and actions defined by a SPA-CAT (Sec.~\ref{sec:spa_cats}) to $\mathbb{R}$. This allows generalization over unseen state-action pairs using the $Q$ values for their abstractions. This section describes our approach for learning SPA-CATs using state-action trajectories collected using any sequential decision making algorithm; our overall algorithm integrating the decision-making process, data collection, and the invocation of SPA-CAT learning phases is discussed in the next section.
 
SPA-CATs are learned autonomously through a process of hierarchical refinement. The SPA-CAT $\Delta$ is initialized with the universal abstraction where $\Delta$ has a single node corresponding to the entire state space, and each action's APT associated with this node has a single node capturing  that action's entire parameter space. 
% A key contribution of this work is a novel algorithm for learning to flexibly refine abstractions in a structured and data-driven manner. Specifically, our method focuses on refining state abstractions by selectively expanding the abstraction hierarchy.
The refinement process creates children nodes for nodes at the fringes of the SPA-CAT and at the fringes of the APTs associated with SPA-CAT nodes.  
These refinements increase the granularity of abstraction in regions of the state and action spaces where finer distinctions are necessary for high performance decision-making.

% Prior work~\citep{dadvar2023conditional} considers a restrictive uniform binary refinement strategy in which abstract states are refined by bisecting each variable range. 
% using statistical learning techniques. 

Suppose the RL agent encounters a set of execution traces of the form $\gD = \{ \langle s_0, a_0, r_0, \dots s_n,a_n,r_n \rangle \}$ where $s_i$ is a concrete state, $a_i$ is an action executed in $s_i$, and $r_i$ is the incurred reward for the transition. These traces are abstracted using the current version of $\Delta$ to produce 
$\bar{\gD} = \{ \langle \overline{s_0}_{\Delta}, \overline{a_0}_{s_0, \Delta},
\bar{r}_0,\dots,\overline{s_n}_{\Delta},\overline{a_m}_{s_m, \Delta},\bar{r}_m \rangle\}$. The abstract sequence is constructed to avoid consecutive duplicate abstract states: trajectory subsequences  $\langle s_i, a_i, r_i, \dots s_{i+k},a_{i+k},r_{i+k} \rangle$  whose state-action segments are abstracted to the same pair  are represented only once as  $\langle \overline{s_i}_\Delta, \overline{a_i}_{s_i,\Delta}, \bar{r}_i\rangle$, where $\bar{r}_i$ is the total cumulative discounted reward for the original subsequence. 
 % We define a function $\gD(\abss,\absa)$ that return a set of partial traces such that $\{ \langle s_0,  a_0, r_0, \dots, s_j, a_j, r_j \rangle \}$ such that $\forall s_i, s_i \in \abss$ and $\forall a_i, a_i \in \absa$. Similarly, we define additional functions $\gD(\abss)$, $\gD(\absa)$, $\gD(s)$, and $\gD(a)$ to represent similar datasets.  

The learning process aims to create agglomerative abstractions where regions of the state and action space that portend similar futures are grouped together. This indicates that dispersions in the value function estimates of abstract states could be used to identify areas where heterogeneous states are incorrectly being combined into an abstraction. However, during early stages of learning, the agent's policy can vary significantly, and the paucity of data makes value-function estimates  extremely unreliable as indicators of similarity in future courses of action.
To balance these considerations, we define a novel hybrid formulation of heterogeneity to identify elements of the current state-action abstraction that need to be refined.

Given  abstract traces $\bar{\gD}$, the TD error $\delta(\bar{s}_i, \bar{a}_i)$ for each subsequent abstract state and action is defined as follows:

\begin{equation}
\begin{aligned}
    \delta(\abss_i,\absa_i) &= (\bar{r}_i + \gamma\max_{\absa} \overline{Q}(\abss_{i+1}, \absa)) - \overline{Q}(\abss_i, \absa_{i}) 
    % SD\left[{\delta}(\abss_i,\absa_i)\right] &= \max_{\absa}\left[\frac{1}{n} \sum_{i=1}^n \delta(\abss_i, \absa_i)^2 - \left( \frac{1}{n} \sum_{i=1}^n \delta(\abss_i, \absa_i) \right)^2 \right]   \\ 
    \label{eq:1}
\end{aligned}
\end{equation}

It is well-known that value-function is inaccurate at start of Q-learning, thus unsuitable for early refinement, whereas, TD-error better represents similar futures during early learning (e.g., \cite{kearns1998finite}). Thus, in early stages of learning, we rely on 
the value of the temporal difference (TD) error to provide a stronger signal of behavioral inconsistency: if $\delta(\bar{s}, \bar{a})$ values show a high standard deviation ($SD$) for an abstract state-action pair in $\bar{\gD}$, then this abstract pair may be representing heterogeneous regions where the $Q$ function is changing at significantly different rates.  On the other hand, as learning progresses and the policy stabilizes, value function estimates become more reliable. At this point the variability of value-function estimates across concrete states in an abstract state are a better indicator of heterogeneity in the abstraction. Therefore, we blend TD error and value-function dispersion metrics. Since it is infeasible to maintain a tabular representation (e.g., a Q-table) over all continuous concrete states and actions, we  compute an estimate of the concrete-state value function as follows: 
\begin{equation}
    \begin{aligned}
      \hat{\delta}(s_i,\absa_i) &= r_i + \gamma \max_{\absa}\overline{Q}(\abss_{i+1}, \absa) - Q(s_i, \absa_i) 
    \\
        \hat{V}(s_i) &= \max_{\absa} [ Q(s_i, \absa) + \alpha * \hat{\delta}(s_i,\absa) ]
    \end{aligned}
    \label{eq:2}
\end{equation}

This allows us to estimate the value function for a state  using a learned Q-value function for abstract states and abstract actions.
We capture the dispersion of $\hat{V}$ estimates across all $n$ concrete states $s_i$ present in the dataset $\gD$. 
% \begin{equation}
%     \begin{aligned}
%         \sigma_{\bar{s}}\left[\hat{V}\right] = \frac{1}{n} \sum_{s_i \in \abss} \hat{V}(s_i)^2 - \left( \frac{1}{n} \sum_{s_i \in \abss} \hat{V}(s_i) \right)^2
%     \end{aligned}
% \end{equation}

% \textcolor{red}{check heterogeneity} 
We combine the standard deviation over TD errors and over $V$ function estimates into a novel heterogeneity estimate for state-action pairs in $\bar{\gD}$:
\begin{equation}
    \begin{aligned}
        H(\abss_i,\absa_i) = \beta \cdot SD_{\bar{\gD}} \left[\delta(\bar{s}_i, \bar{a}_i)\right]  
        + \\ (1-\beta) \cdot SD_{\gD}\left[\hat{V}(s_i) \right]_{s_{i} \in \abss_{i}}
        % \sigma_{\bar{s}_i}\left[\hat{V}\right]
\label{eq:refine_utility}
    \end{aligned}
\end{equation}
Here, the standard deviation is computed over all occurrences of the pair $\bar{s}_i, \bar{a}_i$ in $\bar{\gD}$. 
A scheduling mechanism is used to gradually shift emphasis from TD error  to value function dispersions. This is achieved with a weighting parameter \( \beta \), initialized at 1.0 and annealed across episodes by a decay schedule. States with high heterogeneity, under the current $\beta$, are selected for refinement into finer abstractions.

% We compute variability of a value function using the estimated value for each concrete state. This is needed as the concrete states are continuous and we do not (can not) maintain a Q-table for concrete states. Therefore, variability in the value function is computed as follows:

We rank each abstract state-action pair using the computed heterogeneity $H$ and select top-$k$ abstract states and abstract actions to refine. We use $H(\abss) = \max_{\absa}H(\abss,\absa)$ to select abstract states for refinement and use $H(\abss,\absa)$ for selecting abstract actions for refinements.

    \begin{figure}
        \centering
        \includegraphics[width=0.95\linewidth]{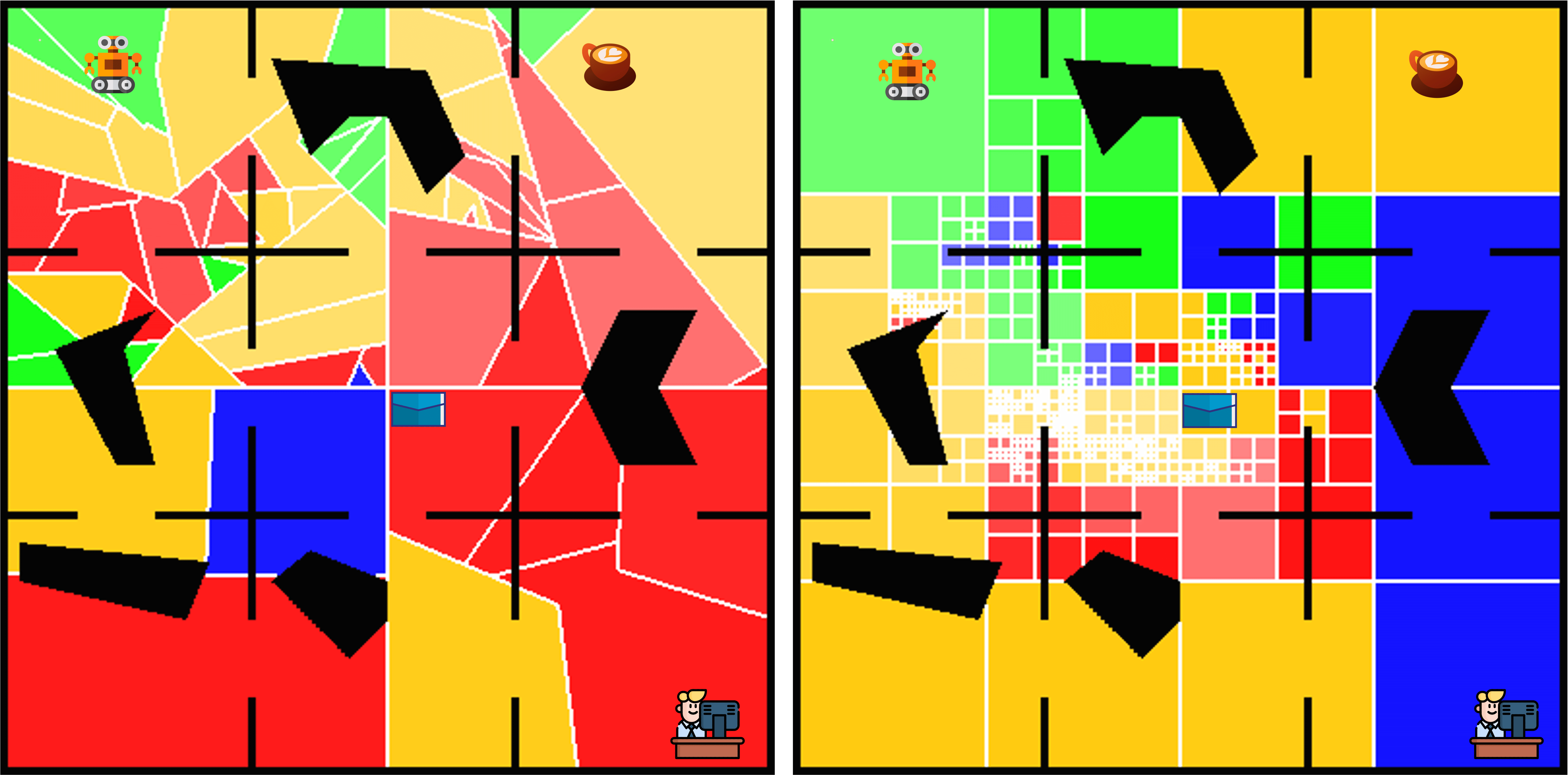}
        \caption{Learned state abstractions using flexible (left) and uniform (right) refinement strategies. The agent is at the top left; it must deliver both coffee and mail to the bottom right. Black lines and regions indicate obstacles. Colors represent actions (yellow: right, green: down, red: up, blue: left).}
        \label{fig:visualization}
    \end{figure}

Multiple paradigms can be used for refining the abstract state-action regions that feature a high heterogeneity under this formulation. We consider two paradigms: uniform refinement as proposed in prior work~\cite{dadvar2023conditional}, and a novel flexible refinement that uses statistical learning. Both state abstractions (nodes for SPA-CATs) and action abstractions (nodes for Action Trees) can be refined using these methods. However, for brevity, we describe them in the context of refining state abstractions. 

\paragraph{Uniform refinement} Uniform partitioning refines an abstract state $\abss$ by independently bisecting each variable's interval to create an orthogonal binary tree decomposition of the state space (see Fig.\,\ref{fig:visualization} (right)). While straightforward, such abstractions are best suited for domains where the Q-function can be factorized into functions over individual state variables and they require extensive refinements to express regions that feature homogeneous value function estimates, but do not constitute hypercubes. 

\paragraph{Learning flexible refinements}  We introduce a novel learning-based approach for constructing flexible refinements. Given an abstract state $\abss$ selected for refinement and the associated set of execution traces, we partition $\abss$ into at most $\gK$ finer abstract states by clustering the concrete states contained within $\abss$.  Specifically, we apply Agglomerative Clustering~\citep{murtagh2012algorithms} from scikit-learn \cite{pedregosa2011scikit} with an adaptive distance threshold: starting from 0.1, we incrementally increase the threshold by 0.001 until the number of clusters is below a specified maximum. This prevents over-fragmentation while ensuring meaningful behavioral distinctions are captured. We use the following similarity criterion to form coherent partitions that reflect underlying behavioral distinctions: 
\begin{equation}
    \begin{aligned}
        J(s) &= \beta \cdot \hat{\delta}(s) + (1-\beta) \cdot \hat{V}(s) \\ 
        \hat{\delta}(s_i) &=  r(s_i, \absa) + \gamma \max_{\absa'} \overline{Q}(\abss_{i+1},\absa') - Q(s_i, \absa_i), 
        \\ \text{where } 
        \absa &= \argmax_{\absa}H(\abss,\absa)
    \end{aligned}
    \label{eq:similarity}
% \begin{equation}
%     \begin{aligned}
%         J(s) &= \beta \cdot \hat{\delta}(s) + (1-\beta) \cdot \hat{V}(s) \\ 
%         \hat{\delta}(s_i) &= \frac{1}{n} \sum_{a_i} r(s_i, a_i) + \gamma \max_{\absa} \overline{Q}(\abss_{i+1},\absa) - \overline{Q}(\abss_i, \absa_i) 
%     \end{aligned}
\end{equation}
Here, $\hat{\delta}$ and $\hat{V}$ are estimated TD-errors and state values for a concrete state, and $\absa$ is the abstract action with high heterogeneity for an abstract state $\abss$. We use a schedule similar to the heterogeneity estimate shift to prioritize using TD-errors estimates earlier in the learning and state-values later in the learning using an annealed parameter $\beta$. 

% To compute clusters, we begin with a small distance threshold of 0.1 and incrementally increase it by 0.001, using agglomerative clustering \cite{pedregosa2011scikit}, until the number of clusters falls below a specified maximum. This approach helps prevent the number of clusters from becoming excessively large.

Once the partitions are identified, we train an SVM classifier to learn decision boundaries between them and define abstract states, with each partition corresponding to a new abstract state. We use balanced class weights and select the regularization parameter through cross-validation based on the smallest class size, evaluating both RBF and linear kernels. This yields refined abstract states that more effectively capture variations in decision-relevant signals like TD error and value estimates, enabling more expressive abstractions.

We now discuss our algorithm for autonomously learning a SPA-CAT and a policy for a given problem.

\begin{algorithm}[t] 
    \caption{\alg{}}
    \label{alg:main}
    \DontPrintSemicolon
    
    \KwIn{MDP $\gM = \langle \gV, \gS, \gA, T, R, \gamma, h \rangle$} 
     \KwOut{Policy $\pi$ for MDP $\gM$ and \spacatname{} $\spacat{}$}
     
      Initialize \spacatname{} $\spacat{}$ and Qtable $\overline{Q}$ \;
     Initialize buffers $D_{\abss,\absa}$ and $D_{s,\absa}$ \;
     
     \For{episode $=1:n_{epi}$}{
     \tcp{Learning phase} 
     $s \leftarrow \text{reset}()$ \;
    % $\gD \gets \emptyset$  \; 
     \For{$\text{step} =1:h$}{
      $\absa \leftarrow \pi(\overline{Q},\abss{})$ 
      \;
      $\abss', \overline{r}, \{s_i, \bar{a}_i, r_i, \dots,s_k \}$ $\gets \text{execute}(s,\absa)$ \;
             $\gD_{\abss,\absa}\text{.add}(\{\abss{}, \absa, \overline{r}, \abss{}'\})$ \;
       $\overline{Q} \leftarrow \text{updateQvalue}(\abss{}, \absa, \overline{r}, \abss{}')$ \; 
             $\gD_{s,\absa}\text{.add}(\{s_i, \absa_i, r_i, \dots, s_k \}$) \;
       $V \leftarrow \text{updateValue}(s, \absa, r, s', \abss{}')$ \; 
       % $\gD_{s,\absa}\text{.add}(s, \absa, s', r)$ \;
      }
      \tcp{Refinement phase} 
      \If{episode $\bmod$ $n_{\text{refine}} = 0$}{ 
     $\gD_{\abss,\absa} \gets$  computeHeterogeneity$(\overline{Q}, \gD_{\abss,\absa})$  \;
    $\bar{\gS}_\emph{ref}, \bar{\gA}_\emph{ref} \gets$ findImprecise($\gD_{\abss,\absa}$)  \;

     \If{refinement == flexible}{
     $\gD_{s,\absa} \gets \text{estimateSimilarity}(V, \gD_{s,\absa}, \bar{\gS}_\emph{ref}, \bar{\gA}_\emph{ref})$ \; 
     $\gC \gets \text{cluster}(\bar{\gS}_\emph{ref}, \gD_{s,\absa}$) \; 
     $\spacat{} \gets$ refine($\gC$, $\bar{\gS}_\emph{ref}$, $\bar{\gA}_\emph{ref}$) \; 
     }
     \Else{
     $\spacat{} \gets $ refine($\bar{\gS}_\emph{ref}, \bar{\gA}_\emph{ref}$) 
     }
      Reinitialize $\gD_{\abss,\absa}$ and $\gD_{s,\absa}$ \;
      }
     }
    \KwRet{$\pi$, $\spacat{}$}
\end{algorithm}

\subsection{\alg{} Algorithm}
\label{sec:pearl}

Alg.~\ref{alg:main} (Parameterized Extended state/action Abstractions for RL, \alg{}), outlines the overall process of how SPA-CAT learning is integrated with TD($\lambda$).
It starts from an initial, coarse \spacatname{} with a single node $N_0$ and one APT for each action $a \in \gA$ with one node each (line 1). It then lets the agent execute in the environment while collecting trajectories and incrementally refining the SPA-CAT.
This enables
\alg{} to jointly learn a \spacatname{} and a policy for the MDP $\gM$. 
It alternates between two main phases: (a) a learning phase (lines 4-11) that trains the policy with a fixed SPA-CAT $\Delta$, and (b) a refinement phase (lines 12-20) that improves the abstraction by refining the \spacatname{}. 
% We now discuss these two phases in detail.

\paragraph{Learning phase}  
In this phase, the agent learns an abstract policy \( \pi : \absS \rightarrow \absA \) over the current \spacatname{} structure using tabular TD-\(\lambda\)~\citep{sutton1988learning} for \( n_{\emph{refine}} \) episodes (lines 4--11). During each episode, the agent follows the abstract policy by executing the corresponding abstract action in the current abstract state, continuing until it reaches a new abstract state or the episode terminates (lines 6--7).

%%%THIS is BUGGY
% Throughout learning, \alg{} collects execution traces of the form \( \gD = \langle s_0, a_0, r_i, \dots, s_n, a_n, r_n \rangle \), where \( s_i \) is the concrete state, \( \absa_i \) is the abstract action taken, and \( r_i \) is the reward received (line 9).
Traces obtained during execution 
$\gD_{\abss,\absa}$ are used to update Q-values and TD errors over abstract state-action pairs using standard TD(\(\lambda\)) updates (Eq.~\ref{eq:1}) (lines 8--9), enabling  policy improvement in the abstract state space. Moreover, traces over concrete state and abstract action pairs $\gD_{s,\absa}$ are used to approximate values of concrete states (Eq.~\ref{eq:2}) (lines 10--11).

\paragraph{Refinement phase}  After every $n_{\emph{refine}}$ episodes,  \alg{} enters the refinement phase to update the \spacatname{} (lines 12-21) via  heterogeneity and similarity measures computed using the methods presented in Sec.~\ref{sec:refinement} (Eq.~\ref{eq:refine_utility} and Eq.~\ref{eq:similarity}). 
%After the refinement is complete, 
The \spacatname{} is then used to continue the learning phase.
%, where \alg{} continues to search for a policy using the refined abstraction until a solution is found.

We now discuss thorough empirical evaluation of our approach in a variety of settings with parameterized actions.

\section{Empirical Results}
\label{sec:results}
We implemented \alg{} along with the annealed heterogeneity estimation and abstraction refinement paradigm presented above. This implementation uses a flexible refinement strategy for refining SPA-CATs and a uniform refinement strategy for refining APTs.
We evaluate \alg{} along three key dimensions:
(1) improvements in sample-efficiency, (2) the quality of the learned policies, and (3) the size of the abstractions generated. Our evaluation is conducted across four challenging SOTA RL domains with stochastic and unknown action models, continuous states, parameterized actions, and sparse rewards (a positive reward only upon reaching the goal). 
Combined with long-horizons, these tasks represent significant challenges for RL. 
% Additional details 
% on test environments
% with their illustrations and 
% additional evaluations, analysis, and 
  % and hyperparameter settings for all methods 
  % are provided in the extended version. 
 % Sec.~\ref{sec:results_supp}.
% The source code and instructions for reproducing our results are also included.

\begin{figure}[t]
    \centering
     \begin{subfigure}{0.15\textwidth}\includegraphics[width=\linewidth]{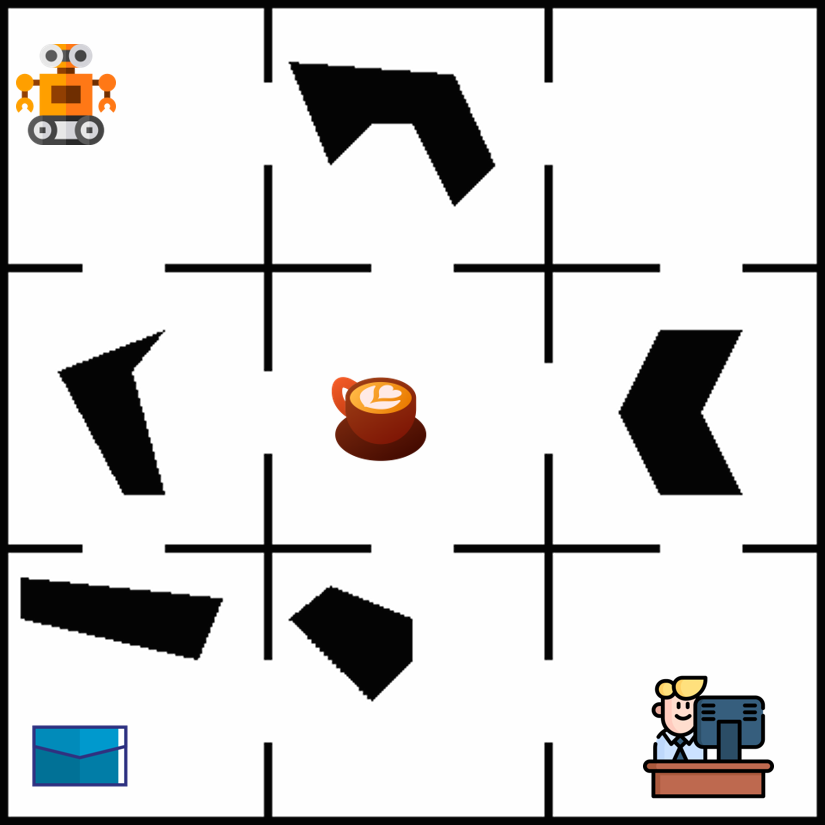}
        \caption{Office}
        \label{fig:sub0}
    \end{subfigure}
    \begin{subfigure}{0.15\textwidth}\includegraphics[width=\linewidth]{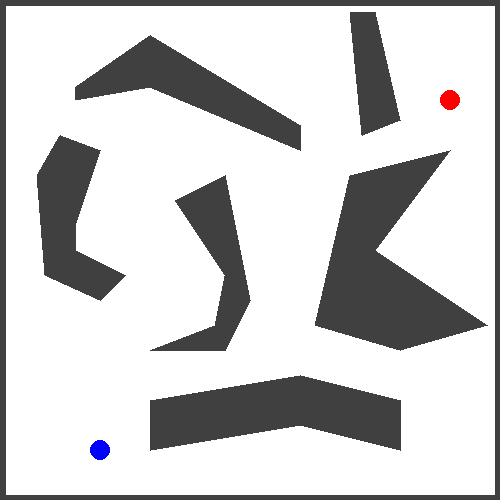}
        \caption{Pinball}
        \label{fig:sub1}
    \end{subfigure}
         \begin{subfigure}{0.15\textwidth}
\includegraphics[width=\linewidth]{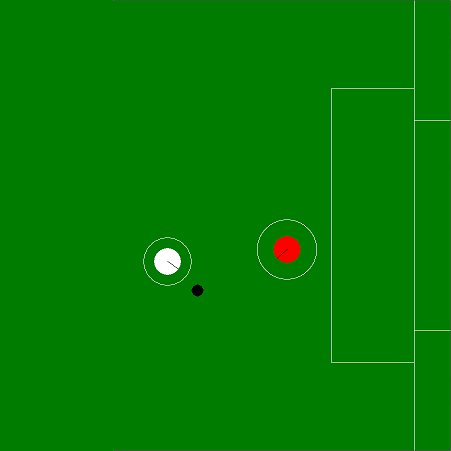}
        \caption{Soccer Goal}
        \label{fig:sub3}
    \end{subfigure}
    % \hfill
    \begin{subfigure}{0.47\textwidth}
\includegraphics[width=\linewidth]{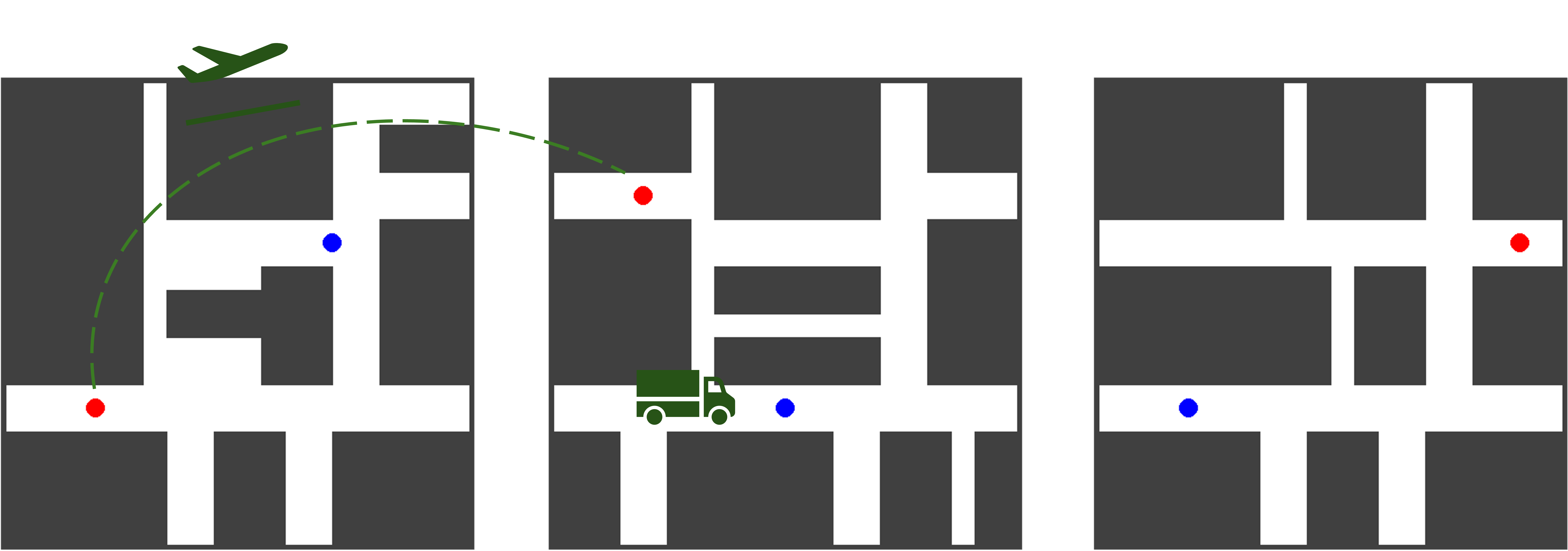}
        \caption{Multi-City Transport}
        \label{fig:sub2}
    \end{subfigure}
    \caption{(a) Office World: The robot needs to pickup coffee and mail and deliver to the office.
    (b) Pinball: A small, dynamic ball needs to be manouvered into a red hole, avoiding collisions with irregularly shaped obstacles.  (c) Soccer Goal: The white agent needs to kick the small black ball past the red keeper. (d) Multi-City Transport: The agent needs to collect a package from a designated location (marked by blue) in a city and deliver to a target airport (marked by red) in a different city. Cities are connected only via airports.}
    \label{fig:envs}
    \vspace{-10pt}
\end{figure}

\paragraph{Test environments} We evaluate on domains well-established as challenging (illustrated in Fig. ~\ref{fig:envs}): (i) OfficeWorld~\citep{icarte2022reward,corazza2024expediting}
(ii) Pinball~\citep{roice2024new,rodriguez2024learning},
(iii) Multi-city transport~\citep{ma2021hierarchical,oswald2024large},
and (iv) Robot Soccer Goal~\citep{bester2019multi}.
Among these, former three are especially challenging due to longer effective planning horizons.

\begin{figure*}[t]
    \centering
    \includegraphics[width=0.96\textwidth]{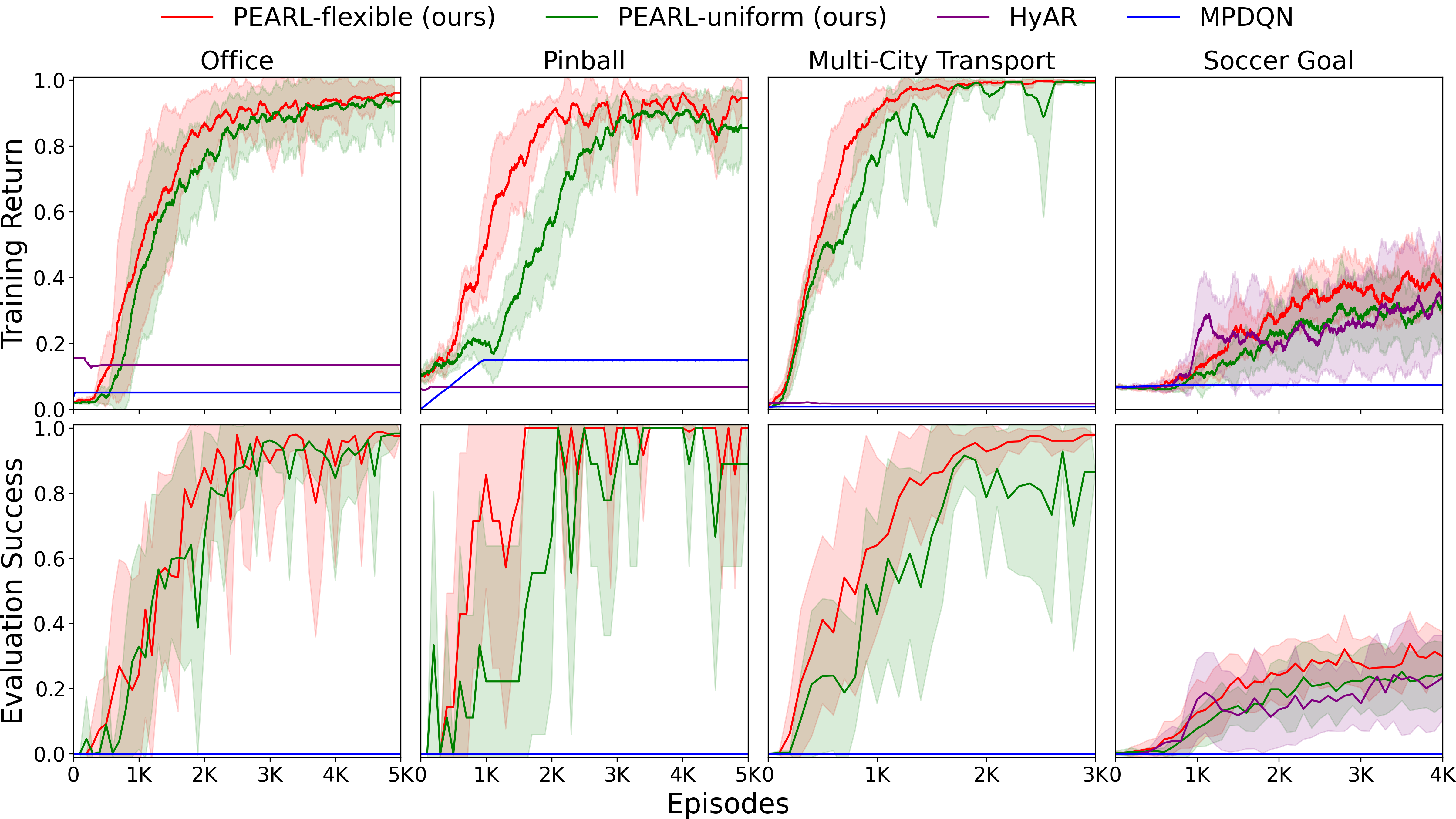}
    \caption{
        Comparison of \alg{}-flexible and \alg{}-uniform with MP-DQN and HyAR in four domains: Office World, Pinball, Multi-City Transport, and Soccer Goal with mean and standard deviation across 50 independent trials.}
    \label{fig:result_supp}
\end{figure*}

\paragraph{Baseline selection} Standard RL approaches---tabular RL~\citep{sutton1988learning,watkins1989learning}, deep RL~\citep{mnih2015human,lillicrap2015continuous,schulman2017proximal,haarnoja2018soft}, hierarchical RL~\citep{nachum2018data, levy2017learning}---are not designed to handle parameterized actions, making them unsuitable as baselines. 
We therefore compare \alg{} against two baselines that support parameterized actions: $(i)$ MP-DQN \citep{bester2019multi}, which extends P-DQN \citep{xiong2018parametrized} by combining DQN and DDPG while addressing P-DQN's over-parameterization problem through multi-pass processing, and $(ii)$ HyAR \citep{li2021hyar}, which learns latent space of hybrid action space and models dependencies between discrete action and continuous parameter using an embedding table and a conditional Variational Auto-Encoder (VAE). 
% Both of these baselines use additional inputs in the form of hand-coded features, which were provided as supplied in their original papers.
To evaluate and compare learning performance fairly without manually biasing learning with head-starts towards favorable solutions, we used the original source code for these baselines while removing their hand-crafted, environment-specific weight initializations. We replaced 
% the domain-engineered initializations 
them
with zero or randomized initializations, whichever yielded better performance. 
% For the baselines, we use publicly available source code, however, we replace the original domain-specific, hand-tuned weight initializations with random initialization.
This provides a consistent, unbiased evaluation of each method's true learning capability.

\paragraph{Metrics and hyperparameters}
We evaluate all agents using two key metrics: (i) cumulative average return during training, and (ii) the success rate of the learned greedy policy. We evaluate two variants of \alg{}---\alg{}-flexible and \alg{}-uniform---which differ in their approach to learning abstractions. 
Reported performance of \alg{} variants include all episodic interactions used for learning state and action abstractions. The results are averaged over 50 independent runs, with both mean and standard deviation reported. 
Full hyperparameter details for all methods are provided in the extended version.

\subsection{Analysis of the Results}

\paragraph{Sample efficiency and performance} Fig.~\ref{fig:result_supp} shows the performance of all methods, with training episodes on the x-axis and two rows of metrics on the y-axis: cumulative return (during training) and success probability (during evaluation).
% ; and the size of the state abstractions learned by \alg{} variants.  
% Although both MP-DQN and HyAR utilize hand-coded features during learning and 
Despite learning abstractions from scratch, both \alg{} variants consistently outperform the baselines across all domains.
This highlights the effectiveness of jointly learning state and action abstractions during RL. Notably, \alg{}-flexible achieves the highest overall performance, demonstrating the benefits of adaptive refinement over a fixed, uniform strategy. In contrast, HyAR fails to learn effective policies in all but the Soccer domain, while MP-DQN fails across all tasks. Note that our comparison excludes the additional episodes HyAR requires to gather experience for training its continuous action embeddings, making the advantage of \alg{} even more pronounced.

\begin{figure}[t]
    \centering
\includegraphics[width=\columnwidth]{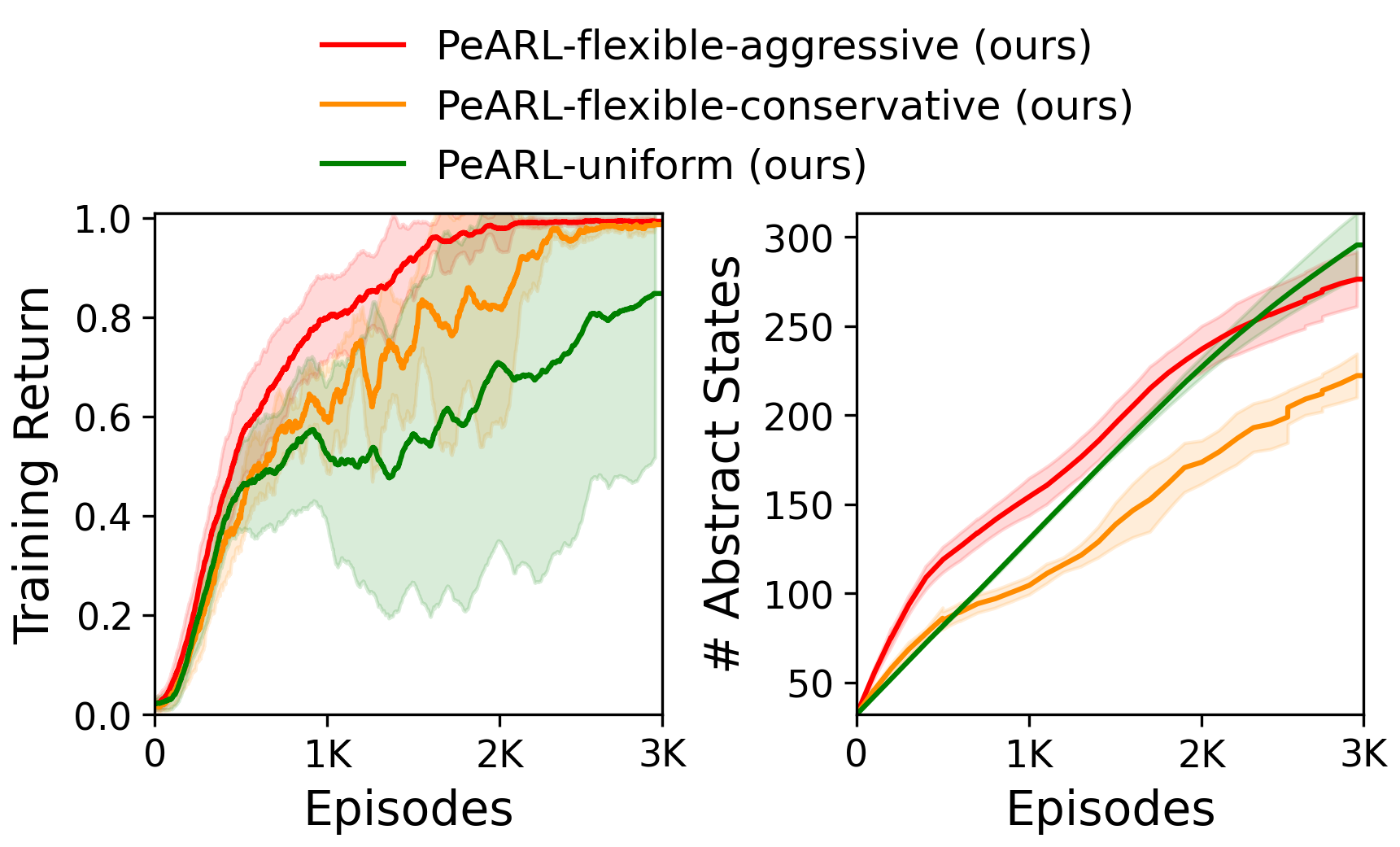}
    \caption{
        Comparison of training reward and state abstraction size for two \alg{}-flexible variants: aggressive and conservative, and \alg{}-uniform in Multi-city Transport.}
    \label{fig:result_supp_second}
\end{figure}

\begin{figure*}[t]
    \centering
\includegraphics[width=0.96\textwidth]{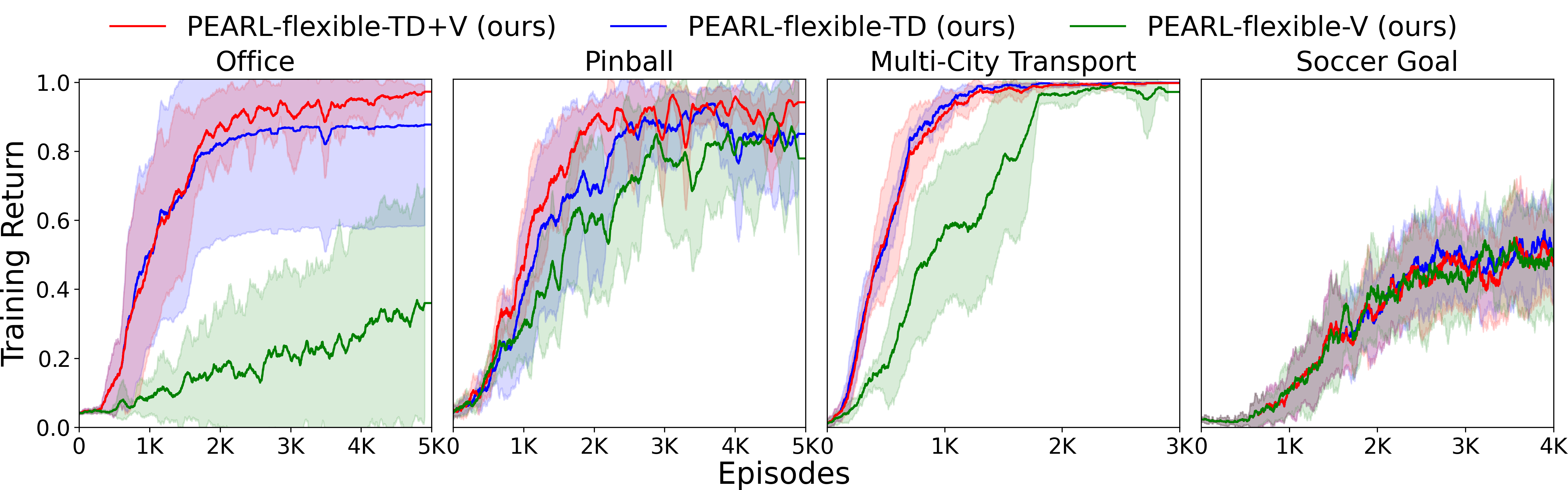}
    \caption{
    Comparison of \alg{}'s performance across all domains under different settings of the annealing hyperparameter $\beta$: TD+V ($\beta$=1.0 with a decay of 0.02 applied at each abstraction refinement step), TD ($\beta$=1.0), and V ($\beta$=0.0).
        }
\label{fig:results_ablation}
\end{figure*}

\paragraph{Parsimony of abstractions} Among the two \alg{} variants, \alg{}-flexible 
offers greater flexibility in controlling abstraction granularity. 
To investigate how abstraction granularity influences learning, we compare three configurations: two \alg{}-flexible variants---aggressive vs. conservative refinement (controlled by varying the maximum number of abstract states allowed for generation per refinement)---alongside \alg{}-uniform. Fig.~\ref{fig:result_supp} shows how these refinement strategies influence both the quality of the learned policies and the size of the resulting abstractions in the Multi-city Transport domain. The aggressively refined \alg{}-flexible variant achieves the highest overall performance, demonstrating the benefits of fine-grained abstractions for precise control. In contrast, the conservatively refined variant achieves comparable performance to \alg{}-uniform but results in a more compact abstraction. These results highlight a key strength of \alg{}-flexible: its ability to adjust the level of abstraction granularity based on task requirements, allowing for better tuning of the trade-off between policy performance and representational simplicity. 

\paragraph{Impact of annealing parameter} Fig.~\ref{fig:results_ablation}
shows that blending dispersion over TD-errors and  values yields better performance than relying on either alone. The annealing hyperparameter provides flexibility to smoothly trade off between these signals during training.

\paragraph{Computational cost} Experiments show that \alg{} has significantly lower runtime than baseline methods since PeARL’s abstraction avoids costly DNN backprops (see the extended version for runtimes).

These results support our hypothesis that jointly learning state-action abstractions improves RL efficiency, enabling TD($\lambda$) to outperform SOTA methods. \alg{}-flexible enables principled control over abstraction granularity, balancing computational simplicity and performance.

\section{Related Work}
\label{sec:related}
% \nsnote{need to expand related work. Also use this to convey the point that most of the existing RL approaches do not work with parameterized action spaces. Especially with continuous parameters.}
\paragraph{Parameterized actions in RL}
Most standard RL methods ~\citep{mnih2015human,lillicrap2015continuous,schulman2017proximal,haarnoja2018soft} are designed for homogeneous action spaces, handling either purely discrete or purely continuous action spaces. Moreover, their success has mostly been limited to settings with short effective horizons, where multi-step lookahead is unnecessary~\citep{laidlaw2023bridging}. 
Parameterized actions, which combine discrete actions with associated continuous parameters, present additional challenges they do not address. Some early approaches, such as
Q-PAMDP
\citep{masson2016reinforcement} alternate between optimizing discrete actions and their continuous parameters. PADDPG \citep{hausknecht2015deep} collapses all action parameters into a single continuous vector. These methods do not exploit the inherent structure of the parameterized actions (the dependency between discrete actions and their associated parameters) essential for learning effective policies. 

P-DQN \citep{xiong2018parametrized} directly handles hybrid action spaces without relaxation or approximation by integrating a DQN (to deal with discrete actions) and a DDPG (to deal with continuous actions). However, this approach treats all action-parameters as a single joint input to the Q-network, which results in dependence of each discrete action's value on all action-parameters, not only those associated with that action. To overcome the over-parameterization problem of P-DQN, MP-DQN \citep{bester2019multi} extend P-DQN with a multiple-pass mechanism,  splitting the action-parameter inputs to the Q-network using several passes. H-PPO \citep{fan2019hybrid} decomposes the action space using parallel sub-actor networks---one for discrete action selection and others for parameter learning---guided by a shared critic. HyAR \citep{li2021hyar} learns a latent representation for hybrid actions via a variational autoencoder, enabling standard DRL algorithms. These methods incur added computational cost due to architectural complexity and hyperparameter sensitivity.

% propose decomposition of action space using multiple parallel sub-actor networks: one for discrete action selection and others for learning their associated parameters, with a shared critic to guide them. HyAR \citep{li2021hyar} learn a decodable representation space for discrete-continuous hybrid actions using a variational autoencoder and then apply conventional DRL algorithms. These approaches introduce additional computational overhead due to architectural complexity and sensitivity to hyperparameters.

% separates actions from action parameters but does not capture the dependence between discrete and continuous components of hybrid actions, required for effective policy learning. 

\paragraph{Abstraction refinement in RL}  
Coarse-to-fine RL (CRL)
\citep{seo2024continuous} discretize continuous action spaces by learning a single action discretization that spans the entire state space. This is achieved by independently learning a Q-network for each action dimension. In contrast, our method learns distinct abstractions of parameterized actions conditioned on abstract states. Unlike prior top-down abstraction methods limited to discrete actions \citep{dadvar2023conditional, nayyar2025autonomous}, \alg{} handles parameterized actions with continuous parameters via action abstraction and supports flexible refinement to compactly capture structure in the problem.

% Unlike previous top-down abstraction-based approaches \citep{dadvar2023conditional, nayyar2025autonomous}, which were restricted to domains with discrete actions, \alg{} addresses parameterized actions through abstraction of actions. \alg{} also supports flexible abstraction refinement, allowing it to capture the underlying structure of the problem space using compact representations.

% \nsnote{cite papers that RLC reviewers suggested..} 

% , while autonomously determining the levels of refinement.

% select sets of intervals for each action dimension, called abstract actions, with highest q-values for refinement. 

% with a high dispersion in TD error values, where refinement on an action dimension depends on o

\section{Conclusion}
We introduced a unified state-action abstraction framework 
with algorithms for learning refinements for an understudied setting of RL with parameterized action spaces. 
Our contributions are: (i) a formalism for context-sensitive abstractions unifying state and action parameters, (ii) a learning-based method for refining state abstractions flexibly, and (iii) \alg{}, an algorithm that jointly learns abstractions during TD($\lambda$). 
% Experiments show \alg{} outperforms existing methods, showing strong sample efficiency and effective policy learning in challenging sparse reward, long horizon domains. 
A theoretical analysis of this framework is a good direction for future work.

\section*{Acknowledgments}
We thank Shivanshu Verma for helping with an earlier version of this approach. This work was supported in part by NSF under grants IIS 2419809 and IIS 1942856.

\bibliography{aaai2026}

\renewcommand{\thesubsection}{\thesection}
\renewcommand\thesection{\Alph{section}}
\setcounter{section}{0}

\section*{Appendix}

\section{Additional Evaluations}
\label{sec:additional_evals}

\paragraph{Computational cost}
Experiments show that \alg{}'s runtime is significantly lower than baselines even though it runs on one CPU and baselines use GPUs. This is true even when accounting for the time required to learn abstractions (Tab.~\ref{tab:runtimes}). This shows that the computational cost of \alg{} is lower since \alg{}'s abstraction avoids costly DNN backprops. 

\paragraph{Impact of annealing parameter on abstraction size}
Tab.~\ref{tab:sizes} reports the size of the final abstractions learned by \alg{}-flexible under different settings of annealing hyperparameter $\beta$ and its decay.  
Using only value-function yields weak early refinement due to zero initialization of the value estimates.  On the other hand, relying on TD-error dispersion early and gradually shifting weight to value dispersion produces substantially stronger refinement.

\begin{table*}[h]
    \centering
      \caption{Runtime comparison across methods}
    \label{tab:runtimes}
    \begin{tabular}{|p{30mm}|p{25mm}|p{25mm}|p{25mm}|p{25mm}|}
        \hline
        Environments & Time (s) by $\newline$ \alg{}-flexible & Time (s) by $\newline$  \alg{}-uniform & Time (s) by $\newline$ MP-DQN & Time (s) by $\newline$ HyAR \\
        % \hline
        % Office & 3411.12 & 3764.67 & $>$ 345600.0 & 442229.95 \\
         \hline
        Office & 3411.12 & 3764.67 & $>$ 345600.0 & $>$ 345600.0 \\
        \hline
        Pinball & 7851.85  & 6558.59 & $>$ 345600.0 & 335715.66 \\
        \hline
        % Multi-City Transport & 2691.96 & 2943.12 & $>$ 345600.0 & 485483.49 \\
         Multi-City Transport & 2691.96 & 2943.12 & $>$ 345600.0 & $>$ 345600.0 \\
        \hline
        Soccer Goal & 1354.66 & 4090.35 & 17844.40 & 155729.82 \\
        \hline
        % rows go here
    \end{tabular}
    
\end{table*}

\begin{table*}[t]
\centering
\caption{Abstraction sizes across domains and \alg{}-flexible variants.}
\label{tab:abstraction_sizes}
\begin{tabular}{lccc}
\toprule
\textbf{Domain} & \textbf{TD-to-V} & \textbf{TD} & \textbf{V} \\
\midrule
Office          &        480       &  475            &    232       \\
Pinball             &     1768          &    1752         &     1102      \\
Multi-City Transport &        254       &    248         &       196    \\
Soccer goal         &       544        &    532         &      521     \\
\bottomrule
\end{tabular}
\label{tab:sizes}
\end{table*}

\section{Test Environments}
\label{sec:results_supp}

% \begin{figure*}[h]
%     \centering
%      \begin{subfigure}{0.3\textwidth}\includegraphics[width=\linewidth]{plots/1-office.png}
%         \caption{Office}
%         \label{fig:sub0}
%     \end{subfigure}
%     \begin{subfigure}{0.3\textwidth}\includegraphics[width=\linewidth]{plots/pinball.png}
%         \caption{Pinball}
%         \label{fig:sub1}
%     \end{subfigure}
%          \begin{subfigure}{0.3\textwidth}
% \includegraphics[width=\linewidth]{plots/goal_domain.png}
%         \caption{Soccer Goal}
%         \label{fig:sub3}
%     \end{subfigure}
%     \hfill
%     \begin{subfigure}{0.8\textwidth}
% \includegraphics[width=\linewidth]{plots/transport.png}
%         \caption{Multi-City Transport}
%         \label{fig:sub2}
%     \end{subfigure}
%     \caption{(a) Office World: The robot needs to pickup coffee and mail and deliver at the office at the bottom right.
%     (b) Pinball: A small, dynamic blue ball that needs to be manouvered into a red hole, avoiding collisions with irregularly shaped obstacles.  (c) Soccer Goal: The white agent needs to kick the small black ball past the red keeper. (d) Multi-City Transport: The agent needs to collect a package from a designated location (marked by blue) in a city and deliver to a target airport (marked by red) in a different city. Cities are connected only via airports.}
%     \label{fig:main}
%     \vspace{-10pt}
% \end{figure*}

% \paragraph{Test environments} 
We evaluate in domains with long-horizons and sparse rewards i.e., agents receive a positive reward only upon reaching the goal state: (i) OfficeWorld~\citep{icarte2022reward,corazza2024expediting} (Fig.~\ref{fig:sub0}): In this domain, the agent must navigate a cluttered indoor office environment to pick up mail and coffee and deliver them to designated office locations. The state space includes the agent’s $(x, y)$ position and two binary variables indicating whether it is carrying coffee or mail. The action space consists of four parameterized movement actions—one for each cardinal direction-with displacement values in the range $[0,0.5)$. The agent automatically picks up items when at their location and drops them off when at the target office location.

(ii) Pinball~\citep{roice2024new,rodriguez2024learning} (Fig.~\ref{fig:sub1}): The agent controls a small ball in a physics-based arena and must guide it into a red hole, avoiding collisions with irregularly shaped obstacles. The ball is subject to dynamic physical forces, such as bouncing off obstacles and surface resistance. The action space includes five parameterized actions: four to increase or decrease velocity in the $x$ or $y$ direction, and one no-op action.

(iii) Multi-city transport~\citep{ma2021hierarchical,oswald2024large} (Fig.~\ref{fig:sub2}): This domain models a complex, multi-city delivery problem. The agent navigates roads within cities and uses air transport to travel between them. The objective is to retrieve a package in one city and deliver it to a destination city. The environment includes three cities, each with an airport. The agent has five parameterized actions: up, down, left, right (each parameterized by distance), and a fly action (parameterized by the destination city), which can only be executed at airports. 

(iv) Robot Soccer Goal ~\citep{bester2019multi} (Fig.~\ref{fig:sub3}): The task involves an agent learning to kick a ball past a keeper. Three actions are available to the agent: kick-to(x,y), shoot-goal-left(y), and shoot-goal-right(y). It terminates if the ball enters the goal, is captured by the keeper, or leaves the play area.

\section{Hyperparameters}
To evaluate and compare the learning performance for all the methods, we use the open-source implementations of MP-DQN\footnote{https://github.com/cycraig/MP-DQN} and HyAR\footnote{https://github.com/TJU-DRL-LAB/self-supervised-rl.git} baselines. However, we replace their manually designed, domain-specific weight initializations with zero or randomized initializations, whichever yielded better results. We retain all other  hyperparameters from their original implementations, as our extensive hyperparameter tuning did not produce better performance than the defaults. 

The hyperparameters used for our methods---\alg{}-flexible and \alg{}-uniform---are detailed in Tables~\ref{tab:tab1} and~\ref{tab:tab2}. Key abstraction-specific parameters include: $\texttt{k\_cap}$ and $\texttt{k\_cap\_actions}$: These define the upper bounds on the number of abstract states and abstract actions, respectively, that are eligible for refinement during each abstraction refinement phase;
$\texttt{max\_clusters}$: Specifies the number of new clusters created when refining an abstract state using flexible refinement, effectively determining how many new abstract states are generated; and
$\texttt{variables\_to\_split}$: Sets the maximum number of state variables considered for uniform refinement at each step. $\texttt{n\_refine}$: Indicates the number of episodes between successive abstraction refinement phases. $\beta$: This is the annealing hyperparameter used for computing the heterogeneity and similarity measures. In addition to these abstraction-related parameters, all standard reinforcement learning hyperparameters (e.g., learning rate, discount factor) are included to ensure reproducibility of experiments. The code, hyperparameters used, and instructions to run experiments are made open-source\footnote{https://github.com/AAIR-lab/PEARL.git}.

% $\texttt{k\_cap}$ and $\texttt{k\_cap\_actions}$ denote the upper bound on the number of abstract states and abstract actions respectively which are selected for refinement of abstraction. $\texttt{max\_clusters}$ denote the number of clusters created when refining an abstract state to generate new abstract states. $\texttt{n\_refine}$ is the interval of episodes after which \alg{} performs an abstraction refinement phase. $\texttt{variables\_to\_split}$ is the upper bound on the number of variables selected for refinement. All the standard hyperparameters of RL are listed.

\begin{table*}[ht]
\centering
\caption{Hyperparameters for \alg{}-flexible used with different domains}
\begin{tabular}{lcccc}
\toprule
% \rowcolor{gray!20}
\textbf{Hyperparameter} & \textbf{Office} & \textbf{Pinball} & \textbf{Multi-City Transport} & \textbf{Soccer Goal} \\
\midrule
\texttt{minimum exploration} $\epsilon_{min}$   & 0.05 & 0.05 & 0.05 & 0.05 \\
\texttt{learning rate}    $\alpha$       & 0.05 & 0.1 & 0.05 & 0.05 \\
\texttt{discount factor} $\gamma$         & 0.99 & 0.999 & 0.99 & 0.99 \\
\texttt{lamda}    $\lambda$        & 0.1  & 0.1  & 0.1  & 0.0  \\
\texttt{maximum steps} $h$        & 400  & 600  & 400  & 150  \\
\texttt{decay}   $\delta$              & 0.9989 & 0.9997 & 0.9989 & 0.9989 \\
\texttt{n\_refine} $n_{refine}$        & 100  & 100  & 100  & 100  \\
\texttt{k\_cap}        & 2    & 40    & 10    & 25    \\
\texttt{k\_cap\_actions}   & 3    & 15    & 10    & 25   \\
\texttt{max\_clusters}     & 3    & 4    & 8    & 20    \\
\texttt{kernel}            & linear & rbf & linear & linear \\
\texttt{annealing $\beta$ decay}            & 0.02 & 0.02 & 0.02 & 0.02 \\
\bottomrule
 \label{tab:tab1}
\end{tabular}
\end{table*}

\begin{table*}[ht]
\centering
\caption{Hyperparameters for \alg{}-uniform used with different domains}
\begin{tabular}{lcccc}
\toprule
% \rowcolor{gray!20}
\textbf{Hyperparameter} & \textbf{Office} & \textbf{Pinball} & \textbf{Multi-City Transport} & \textbf{Soccer Goal} \\
\midrule
\texttt{minimum exploration} $\epsilon_{min}$   & 0.05 & 0.05 & 0.05 & 0.05 \\
\texttt{learning rate}    $\alpha$       & 0.05 & 0.1 & 0.05 & 0.05 \\
\texttt{discount factor} $\gamma$         & 0.99 & 0.999 & 0.99 & 0.99 \\
\texttt{lamda}    $\lambda$        & 0.1  & 0.1  & 0.1  & 0.0  \\
\texttt{maximum steps} $h$        & 400  & 600  & 400  & 150  \\
\texttt{decay}   $\delta$              & 0.9989 & 0.9997 & 0.9989 & 0.9989 \\
\texttt{n\_refine} $n_{refine}$        & 100  & 100  & 100  & 100  \\
\texttt{k\_cap}        & 5    & 40    & 10    & 10    \\
\texttt{k\_cap\_actions}   & 5    & 15    & 10    & 10   \\
% \texttt{max\_clusters}     & 4    & 4    & 4    & 4    \\
% \texttt{kernel}            & linear & rbf & linear & linear \\
\texttt{variables\_to\_split} & 4   & 2    & 4    & 2   \\
\texttt{annealing $\beta$ decay}            & 0.02 & 0.02 & 0.02 & 0.02 \\

\bottomrule
 \label{tab:tab2}
\end{tabular}
\end{table*}

% \bibliography{aaai2026}

% \end{document}

\end{document}